\documentclass[10pt,twocolumn,letterpaper]{article}

\usepackage[accsupp]{axessibility}  

\usepackage{iccv}
\usepackage{times}
\usepackage{epsfig}
\usepackage{graphicx}
\usepackage{amsmath}
\usepackage{amssymb}
\usepackage{booktabs}
\usepackage{xspace}
\usepackage{float}
\usepackage{xspace}
\usepackage{xcolor}
\usepackage{enumitem}
\usepackage{caption}[font={small}]
\usepackage{cite}
\usepackage{wrapfig}
\usepackage{adjustbox}
\usepackage{multirow}
\usepackage{subfigure}

\usepackage[pagebackref=true,breaklinks=true,letterpaper=true,colorlinks,bookmarks=true]{hyperref}

\iccvfinalcopy 

\newcommand{\name}{SynBody\xspace}
\newcommand{\hmname}{SMPL-XL\xspace}
\newcommand{\Scale}{1.2\xspace}
\newcommand{\ScaleSMPL}{2.7\xspace}
\newcommand{\ScaleMotion}{1,187\xspace}
\newcommand{\ScaleSeq}{26,960\xspace}
\newcommand{\cm}{\checkmark\xspace}

\def\iccvPaperID{3878} 

\ificcvfinal\fi

\begin{document}

\title{SynBody: Synthetic Dataset with Layered Human Models for \\ 3D Human Perception and Modeling}

\author{
Zhitao Yang$^{1,*}$ \quad
Zhongang Cai$^{1,2,3,*}$ \quad
Haiyi Mei$^{1,*}$ \quad
Shuai Liu$^{2,*}$ \quad
Zhaoxi Chen$^{3,*}$ \quad
\\
Weiye Xiao$^{1}$ \quad
Yukun Wei$^{1}$ \quad
Zhongfei Qing$^{1}$ \quad
Chen Wei$^{1}$ \quad 
Bo Dai$^{2}$ \quad
Wayne Wu$^{2}$ \quad
\\
Chen Qian$^{1}$ \quad
Dahua Lin$^{4}$ \quad
Ziwei Liu$^{3,\dagger}$ \quad
Lei Yang$^{1,2,\dagger}$ \quad
\\
$^{1}$SenseTime Research \quad 
$^{2}$Shanghai AI Laboratory \quad 
\\
$^{3}$S-Lab, Nanyang Technological University \quad
$^{4}$The Chinese University of Hong Kong \quad
\\
$^{*}$Equal Contribution \quad
$^{\dagger}$Corresponding Author \quad
\\
\url{https://synbody.github.io/}
}


\twocolumn[{
\renewcommand\twocolumn[1][]{#1}
\maketitle 
\begin{center}
    \centering
    \vspace{-5mm}
    \includegraphics[width=0.95\linewidth]{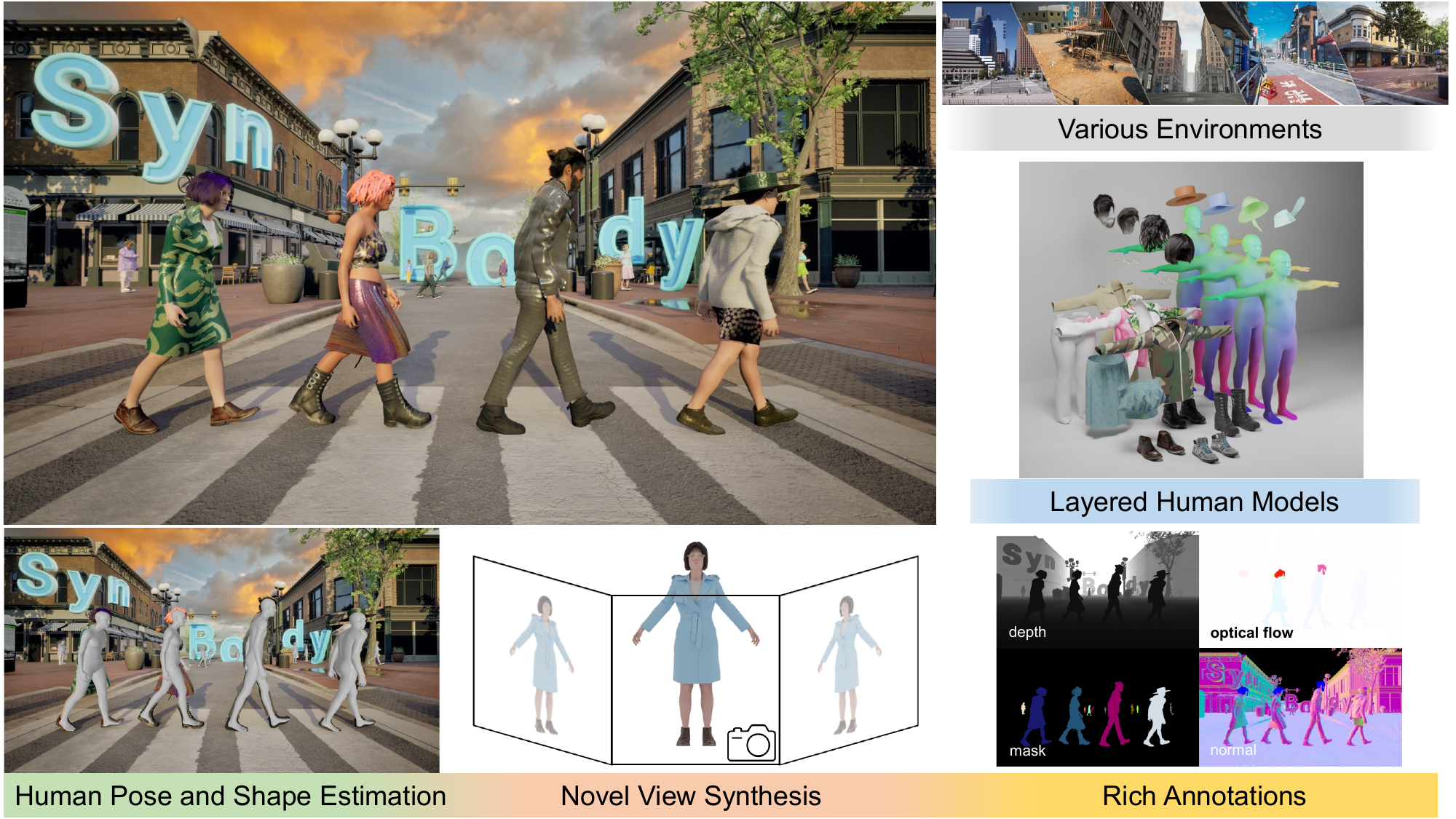}
    \setlength{\abovecaptionskip}{0mm}
    \captionof{figure}{\small
	\textbf{\name is a large-scale synthetic dataset with a massive number of subjects and high-quality annotations.} It supports various research topics, including human pose and shape estimation (HPS) and novel view synthesis for human (Human NeRF). 
	}
	\label{fig:teaser}
\end{center}
}]

\ificcvfinal\thispagestyle{empty}\fi

\begin{abstract}


Synthetic data has emerged as a promising source for 3D human research as it offers low-cost access to large-scale human datasets.
To advance the diversity and annotation quality of human models, we introduce a new synthetic dataset, \textbf{\name}, with three appealing features:
\textbf{1)} a clothed parametric human model that can generate a diverse range of subjects; \textbf{2)} the layered human representation that naturally offers high-quality 3D annotations to support multiple tasks; \textbf{3)} a scalable system for producing realistic data to facilitate real-world tasks.
The dataset comprises {\Scale}M images with corresponding accurate 3D annotations, covering 10,000 human body models, \ScaleMotion actions, and various viewpoints. The dataset includes two subsets for human pose and shape estimation as well as human neural rendering.
Extensive experiments on \name indicate that it substantially enhances both SMPL and SMPL-X estimation. 
Furthermore, the incorporation of layered annotations offers a valuable training resource for investigating the Human Neural Radiance Fields(NeRF). 
\end{abstract}
\section{Introduction}
\label{sec:introduction}


The fields of 3D human perception~\cite{kanazawa2018end,kolotouros2019learning,kocabas2020vibe,kocabas2021pare,zeng2021learning,pang2022benchmarking} and human reconstruction~\cite{peng2021neural,peng2021animatable,kwon2021neural,hong2022eva3d,hong2022avatarclip} have become increasingly important, but the lack of available data has limited their development. Collecting real human data on a large scale is challenging due to privacy concerns and time constraints. Therefore, exploring the use of synthetic human datasets has become a critical avenue of research.

Despite the great potential, existing synthetic datasets~\cite{Varol2017LearningFS,patel2021agora,bazavan2021hspace,cai2021playing} suffer from limitations such as the number of available human models and the quality of annotations.
The main reason lies in that synthetic human datasets rely on real scans for rendering, which poses three key obstacles. Firstly, it is challenging to expand the types of body shapes, poses, and clothing available in the dataset. Secondly, as the human models are scanned with clothing, the 3D annotations obtained through fitting are prone to errors. Thirdly, it is difficult to obtain annotations of body and clothing separately.
To address these issues, we develop a new synthetic dataset termed \name. The dataset includes \Scale million frames with corresponding ground-truth 3D human body annotations. It covers 10,000 human body models, \ScaleMotion motions, and \ScaleSeq video clips with {\ScaleSMPL}M SMPL/SMPL-X annotations.

At the heart of \name is the layered parametric human model, which constructs the clothed human model in a bottom-up manner. SMPL-X~\cite{pavlakos2019expressive} is a widely used parametric human model, capable of sampling human models with various body shapes. However, it lacks the ability to model clothing, limiting its applicability when synthesizing realistic human models.
To overcome this limitation, we introduce \hmname, a parametric human model based on SMPL-X in a layered representation. \hmname enriches the SMPL-X model in three aspects:
(1) Hair system: adding hair and beards to the FLAME~\cite{li2017learning} model, with 32 types of hair and 13 types of beards;
(2) Garment and accessories: adding procedural clothes to the SMPL-X body, including coats, shirts, pants, skirts, shoes, and glasses;
(3) Texture: in addition to adding rich geometry, \hmname also adds rich textures for sampling various skin colors and clothing textures.

The designed \hmname is capable of automatically generating a large number of human models with high-quality annotations.
We therefore generate 10,000 clothed human models by sampling various body shapes, clothing styles, hairstyles, accessories, and textures.
Notably, the use of the SMPL-X model as the base body model guarantees that the parametric human annotations are always accurate, obviating the need for the necessity for annotations through fitting. Furthermore, as the clothes are explicitly attached to the surface of the human body, layered annotations for body and clothes are available.

To generate a large-scale dataset with high diversity and high-quality annotations, we design a scalable and automatic system to render images and annotations.
We first animate the 10,000 dressed human models by retargeting motions from a large motion library~\cite{mahmood2019amass}. Subsequently, we design an algorithm to place human models in the scene without piercing. Multiple cameras are then placed by evaluating self-occlusion, inter-occlusion, and view diversity, and the rendering module renders the assets into images with corresponding annotations.

%
With \name, we launch two tracks that support human pose and shape estimation and human neural rendering, respectively.
Experiments show that \name is more effective than AGORA under the same amount of training data for human pose and shape estimation. With diverse and large-scale training data, \name achieves significant performance gains on both SMPL and SMPL-X estimation.
In terms of human neural rendering using neural radiance fields (\ie NeRF~\cite{mildenhall_nerf_2020}), benchmarking existing approaches on \name shows that it has comparable performance as real human data. Furthermore, with the layered annotations which offer accurate SMPL parameters, we observe that current human NeRF approaches are sensitive to the accuracy of estimated SMPL.

In summary, \name is a large-scale synthetic dataset for human perception and modeling, with three main contributions:
(1) It constructs clothed subjects and samples 10,000 animatable subjects, which is an order of magnitude higher than existing datasets.
(2) The clothed subjects are constructed with an explicit cloth model, thus it provides layered 3D annotations of the human body and clothing, which are not available in previous datasets.
(3) Experiments on \name achieve promising results on both human perception and modeling, emphasizing the importance of diversity and annotation quality for downstream tasks.
\section{Related Works}
\label{sec:related_works}

\begin{table*}[t]
  \begin{center}
  \caption{
  \textbf{Comparisons of 3d human dataset.} 
  We compare \name with existing datasets. We divide datasets into three types: real (R), synthetic (S), and mixed (M). \name constructs 10,000 animatable subjects, which is an order of magnitude higher than any existing datasets and brings competitive scale and diversity.
  ``ITW'' stands for ``In-the-Wild'' in the table.
  }
  \label{tab:dataset_comparison}
  \small
  \begin{tabular}{llccclllll}
    \toprule
    Dataset & Type & ITW & Video & 
    \#Views & \#SMPL & \#Seq & \#Subj. & \#Motions & GT format\\
    
    \midrule
    
    HumanEva \cite{Sigal2009HumanEvaSV} & R & - & \cm &
    4/7 & NA & 7 & 4 & 6 & 3DJ\\
    
    Human3.6M \cite{ionescu2013human3} & R & - & \cm &
    4 & 312K & 839 & 11 & 15 & 3DJ, SMPL\\
    
    MPI-INF-3DHP \cite{mehta2017monocular} & M & \cm & \cm & 
    14 & 96K & 16 & 8 & 8 & 3DJ\\
    
    3DPW \cite{von2018recovering} & R & \cm & \cm &
    1 & 32K & 60 & 18 & * & SMPL\\
    
    Panoptic Studio \cite{Joo2019PanopticSA} & R & - & \cm &
    480 & 736K & 480 & $\sim$100 & * & 3DJ\\
    
    EFT \cite{Joo2020ExemplarFF} & R & \cm & - &
    1 & 129K & NA & Many & NA & SMPL\\

    ZJU-MoCap \cite{peng2021neural} & R & - & \cm &
    21 & 180K & 9 & 9 & 9 & SMPL,mask \\
    
    
    SURREAL~\cite{Varol2017LearningFS} & S & \cm & \cm &
    1 & 6.5M & NA & 145 & 2K & SMPL \\ 
    
    AGORA \cite{patel2021agora} & S & \cm & - & 
    1 & 173K & NA & $>$350 & NA & SMPL, SMPL-X, mask\\
    
    HSPACE~\cite{bazavan2021hspace} & S & \cm & \cm & 
    5 & - & NA & 100$\times$16 & 100 & GHUM/L, mask \\

    GTA-Human~\cite{cai2021playing} & S & \cm & \cm &
    1 & 1.4M & 20K & $>$600 & 20K & SMPL\\

    BEDLAM~\cite{black2023bedlam} & S & \cm & \cm &
    - & 1M & 10K & 271 & 2,311 & SMPL-X, mask\\ 
    
    \textbf{\name} & S & \cm & \cm &
    4 & {\ScaleSMPL}M & 27K & 10,000 & \ScaleMotion & SMPL, SMPL-X, mask \\
    
    \bottomrule
  \end{tabular}
\end{center}
\vspace{-0.5cm}
\end{table*}

\noindent \textbf{Human Parametric Models.}
Several 3D human parametric models, such as SMPL~\cite{loper2015smpl}, SMPL-X~\cite{pavlakos2019expressive}, and GHUM~\cite{xu2020ghum}, have been developed to generate 3D human meshes from parameters that represent the human pose and shape using linear blend skinning. SMPL-X~\cite{pavlakos2019expressive} extends SMPL~\cite{loper2015smpl} by combining FLAME~\cite{li2017learning} and MANO~\cite{romero2022embodied} for the head and hands, respectively, and is trained on a large number of real scans to provide a strong basis for shape variations. However, SMPL-X only produces naked body meshes, and we aim to enhance its realism by building a layered parametric model that includes hair, clothes, and accessories. The proposed model leverages the shape basis of SMPL-X while providing realistic dressed human meshes.

\noindent \textbf{Human Pose and Shape Estimation.}
%
Several methods~\cite{kanazawa2018end,kolotouros2019learning,kocabas2020vibe,kocabas2021pare,pang2022benchmarking,wang2023zolly,zeng2022smoothnet,zeng2022deciwatch,zeng2021learning} have been proposed to estimate 3D human pose and shape parameters.
HMR~\cite{kanazawa2018end} directly regresses these parameters in an end-to-end manner, while SPIN uses an optimization step~\cite{kolotouros2019learning} to guide the learning process towards pseudo-3D labels.
PARE~\cite{kocabas2021pare} employs part attention to tackle occlusion, and VIBE~\cite{kocabas2020vibe} leverages temporal information in videos for SMPL estimation.
Apart from building low-cost data collection solutions~\cite{cai2022humman}, synthetic datasets have become a promising alternative to efficiently scale up data.
SURREAL~\cite{Varol2017LearningFS} renders textured SMPL body models in real-image backgrounds but does not account for cloth geometry, resulting in unrealistic subjects.
%
%
AGORA~\cite{patel2021agora} renders real human scans in a virtual world and provides high-quality synthetic data for image-based approaches.
HSPACE~\cite{bazavan2021hspace} places animated human models in various scenes to provide training data for video-based methods, and increases the variation of human shape via refitting.
GTA-Human~\cite{cai2021playing} captures videos and optimizes corresponding SMPL annotations from video games.
However, the diversity of subjects in current datasets is limited by either body shapes or cloth types.

\noindent \textbf{Expressive Human Pose and Shape Estimation.}
%
As face and hand are also crucial for human perception, some efforts~\cite{rong2021frankmocap,moon2022accurate,zhang2022pymaf} have been made to whole-body human pose and shape estimation.
%
ExPose~\cite{choutas2020monocular} introduces three experts to predict parameters for body, hand, and face, and merge them in a copy-paste strategy.
%
%
%
Hand4Whole~\cite{moon2022accurate} further improves the prediction of wrist and finger poses by leveraging selected hand joint features.
Predicting both hands and faces makes the dataset much more difficult to obtain than just predicting the body. In order to increase the diversity of real data, AGORA~\cite{patel2021agora} provides image-based SMPL-X annotations by fitting SMPL-X on the scanned human models.
While a very recent work, BEDLAM~\cite{black2023bedlam}, also enhances the SMPL-X model with clothing, hair, and changes in skin tones. However, it employs a manual approach for creating garments and utilizes pre-made skin textures, making it challenging to generate a large number of human body models. In contrast, our body model construction process is fully automated, encompassing clothing, hair, textures, and more. This enables easier scalability to generate human body models on a large scale.

\noindent \textbf{Human NeRF.}
NeRF~\cite{mildenhall_nerf_2020} has demonstrated impressive photo-realistic view synthesis by learning implicit fields of density and color.
Yet, human motions are more challenging to learn due to dynamic deformation fields.
NeuralBody~\cite{peng2021neural} incorporates prior from a statistical body template to learn dynamic sequence,
while Animatable NeRF~\cite{peng2021animatable} proposes to reconstruct an animatable human model that generalizes to new poses.
Furthermore, NHP~\cite{kwon2021neural} and KeypointNeRF~\cite{Mihajlovic:KeypointNeRF:ECCV2022} achieve generalizability to unseen identities and poses.
Several datasets have been adapted to study human NeRF.
ZJU-MoCap~\cite{peng2021neural} captures 9 human subjects with 21 synchronized cameras, providing fitted human body model parameters as well as the foreground mask.
Human3.6M~\cite{h36m_pami} collects 11 human subjects with 4 cameras, using a marker-based motion capture system.
A-NeRF~\cite{su2021anerf} generates a synthetic dataset using SURREAL~\cite{Varol2017LearningFS} to study factors that affect visual quality.
Powered by the same toolchain~\cite{xrfeitoria} as \name, SHERF~\cite{SHERF} achieves generalizable Human NeRF model for recovering animatable 3D humans from a single input image, and HumanLiff~\cite{HumanLiff} propose layer-wise 3D human generative model with a unified diffusion process.

While the aforementioned tasks are interrelated and often draw upon similar datasets, as shown in Table~\ref{tab:dataset_comparison}, datasets are limited in two key aspects. Firstly, obtaining real human models is challenging, which restricts the scale of these datasets. Secondly, 3D annotations are typically acquired through optimization, which introduces errors and cannot provide layered annotations.
In contrast, with the designed \hmname model, \name provides 10,000 subjects with diverse body shapes and clothing, along with layered annotations that include accurate SMPL and SMPL-X. Built on top of this layered human model, SynBody comprises two subsets for human mesh recovery and human NeRF, respectively, and features the same 10,000 diverse subjects.
\section{Synthetic Data Generation System}
\label{sec:toolchain}

\begin{figure}[htbp]
  \includegraphics[width=\linewidth]{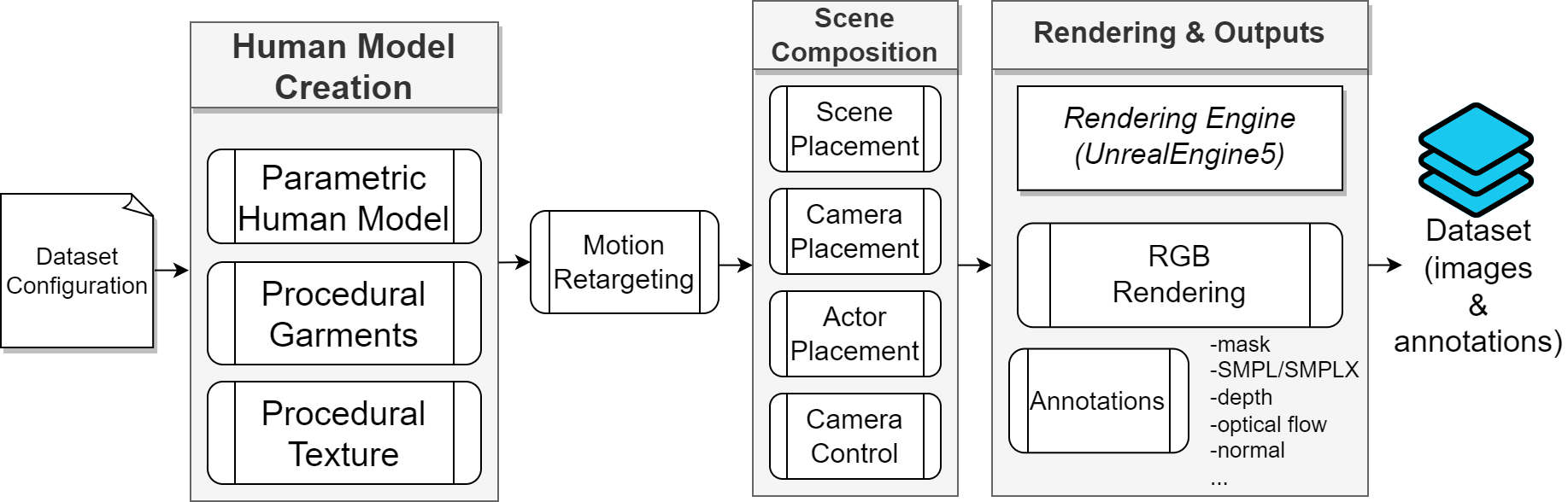}
  \caption{\textbf{Synthetic data generation system.} It consists of 4 components: 1) human model creation to generate layered human models, 2) motion retargeting to drive human models, 3) scene composition to place actors and cameras, and 4) rendering and outputs to generate multi-modal dataset.}
  \label{fig:toolchain}
  \vspace{-0.2cm}
\end{figure}

Similar to movie production~\cite{anyi2023dynamic}, our system comprises 4 components (depicted in Figure~\ref{fig:toolchain}): (1) a layered parametric human model creation service, a scalable process to generate layered human models, (2) a motion retargeting module to apply motions from various sources to layered human models, (3) a scene composition module to place 3D actors and objects into a 3D scene, setting up cameras, (4) a 3D rendering engine and a multi-modal data annotation generator. Our infrastructure enables the generation of high-quality synthetic data for various computer vision tasks.

%
%
%
%
%

\subsection{Layered Parametric Human Model}

Parametric human models like SMPL-X\cite{pavlakos2019expressive} provide the ability of create rigged body models with various body shapes. However, the lack of available textures limits its application in data generation. We designed a module to perform an automatic process that combines SMPL-X with procedural garments and accessories, hair system and textures, producing realistic and diverse body models.


\begin{figure*}
\centering
\begin{minipage}{0.456\textwidth}
  \centering
  \includegraphics[width=\linewidth]{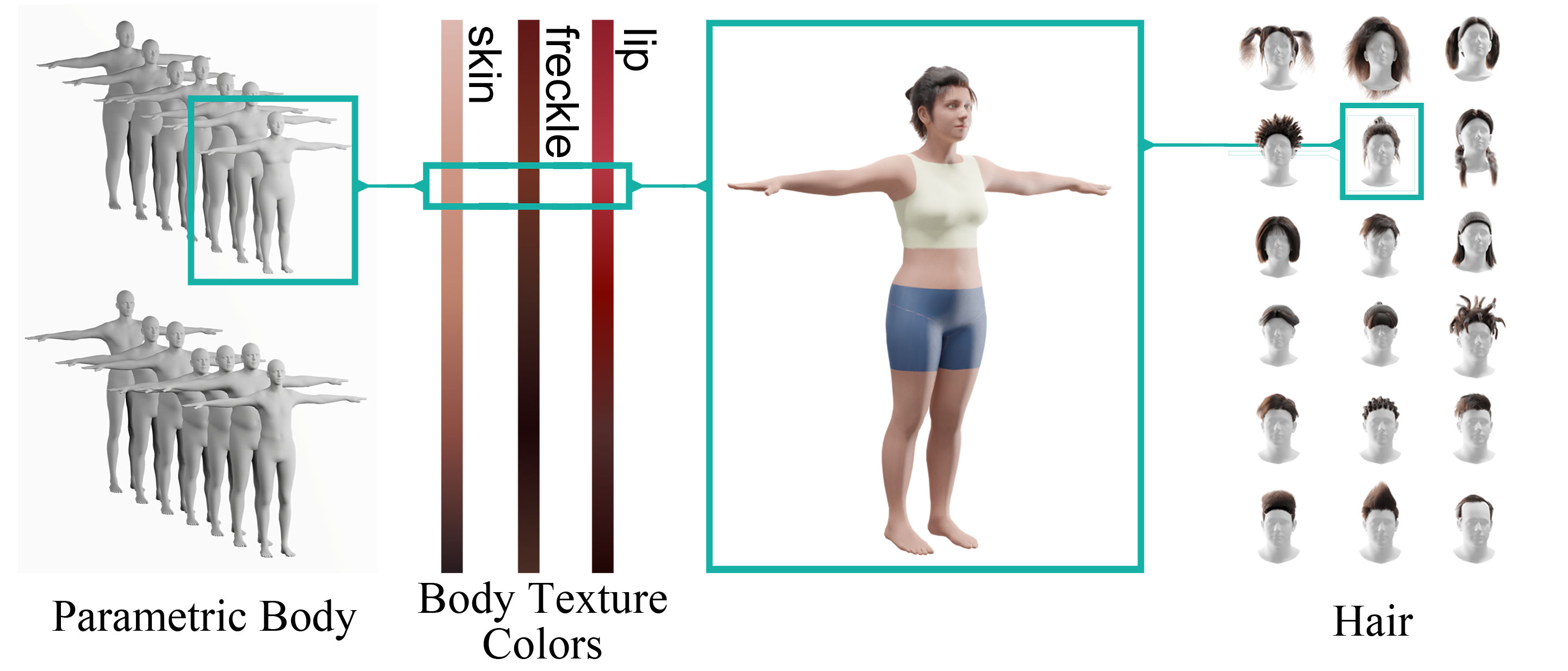}
  \centering
  (a) Naked human model generation
\end{minipage}
\hfill
\begin{minipage}{0.422\textwidth}
  \centering
  \includegraphics[width=\linewidth]{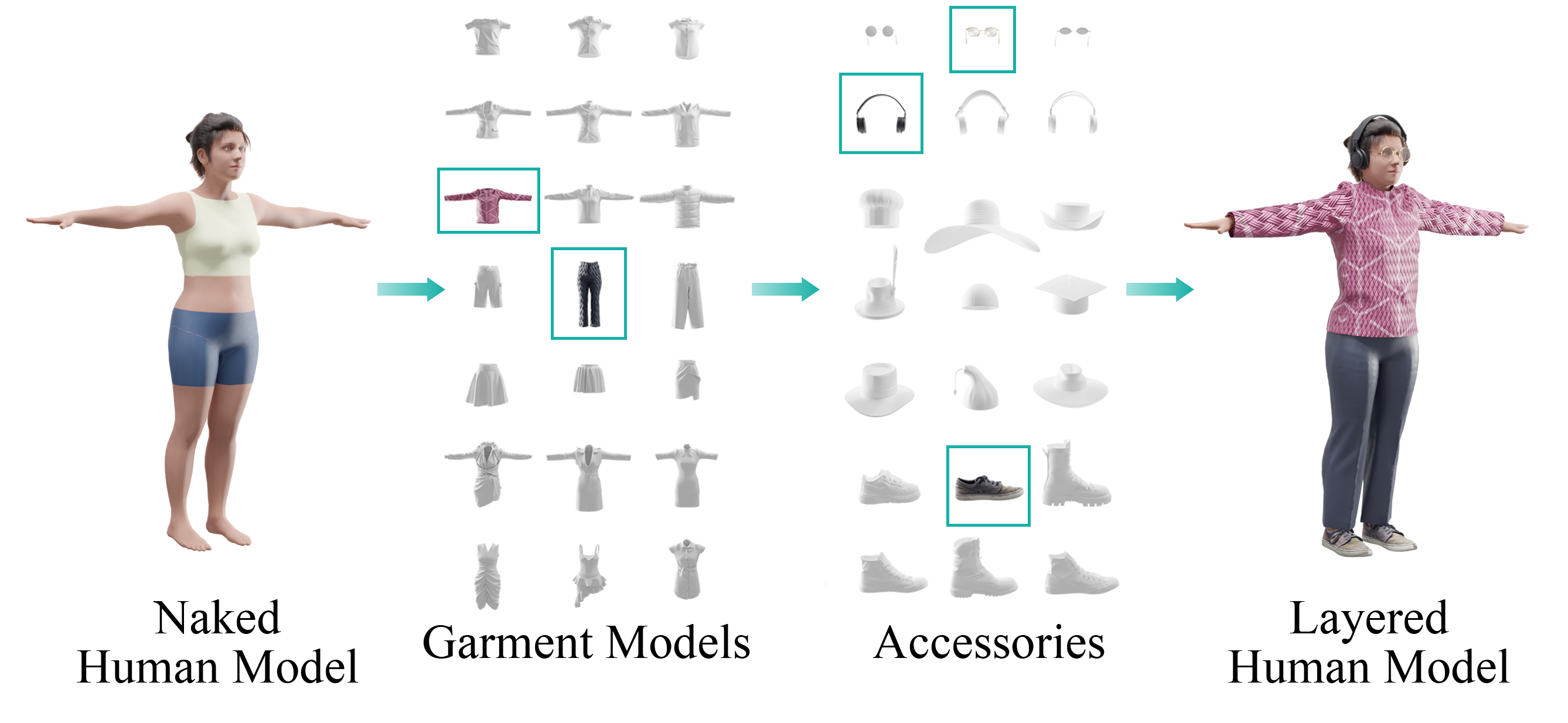}
  \centering
  (b) Layered human model generation
\end{minipage}
\vspace{-2mm}
\caption{Layered parametric human model creation: (a) Generation of a naked human model using parametric body model~\cite{pavlakos2019expressive}, procedural body texture colors, and particle hair System; (b) Integration of garments and accessories onto the naked body model.}
\label{fig:SesneDressing}
\end{figure*}

\noindent\textbf{Body shape.}
%
SMPL-X\cite{pavlakos2019expressive} body model has a 3D mesh whose vertex locations are controlled by parameters for pose $\theta$, shape $\beta$, and facial expression $\psi$. By modifying shape $\beta$, we obtain 3D meshes of various human heights and weights. And by alternating pose $\theta$, meshes can be driven to perform various poses.

\begin{figure}[htbp]
  \includegraphics[width=\linewidth]{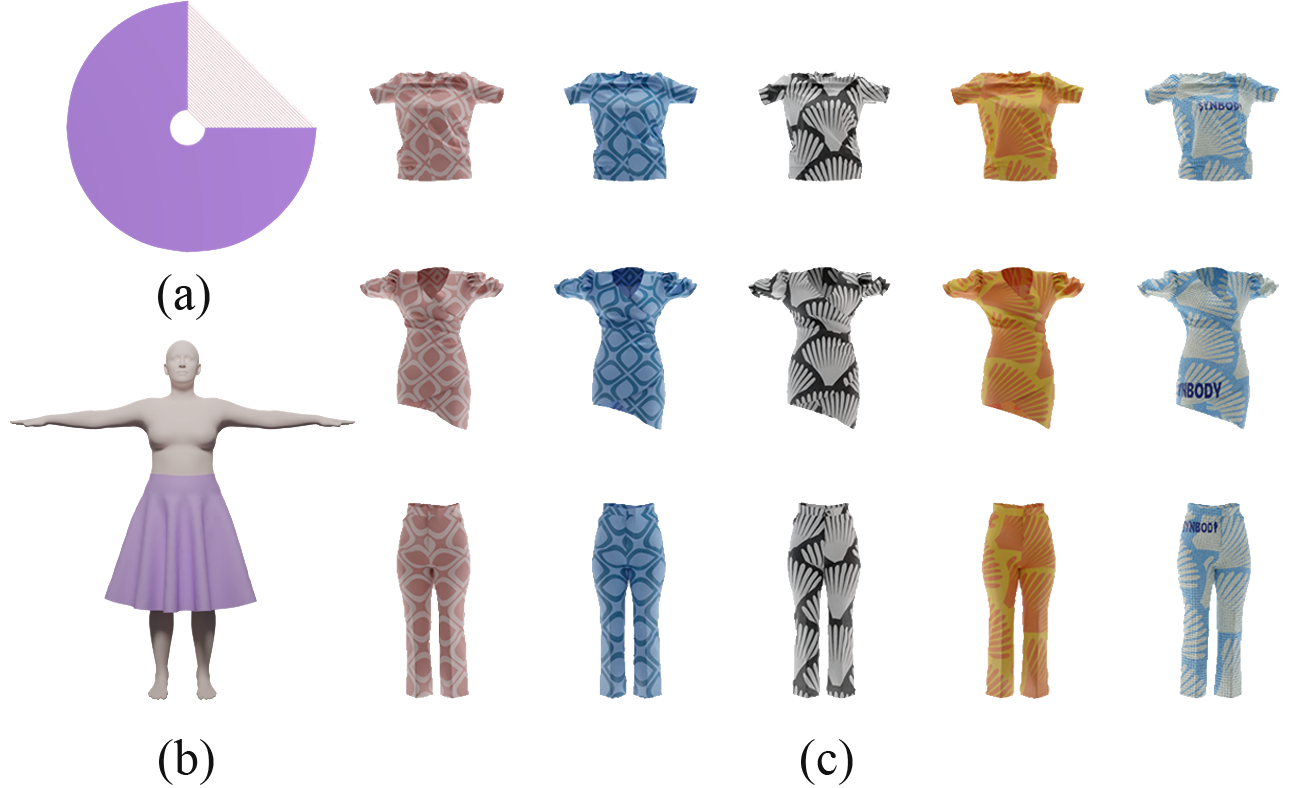}
  \vspace{-4mm}
  \caption{
  Demonstrations of Garment and Procedural Texture. (a) garment pattern with sewing lines, (b) simulated garment under the body in canonical space, (c) procedural textured garment whose style and color are elevated by layered masks provided by patterns and decals.
  }
  \label{fig:ProceduralTexturing}
  \vspace{-2mm}
\end{figure}

\noindent\textbf{Garment Model.}
%
Our garment is generated as a separate layer on top of the body. Following the industrial garment-making workflow, we designed garment patterns of various styles. Notice that different parts of garment pieces are connected with sewing lines, e.g., the red line in Figure~\ref{fig:ProceduralTexturing} (a). Then, we stitch patterns onto the body at T-Pose utilizing a physical simulator\cite{MarvelousDesigner}. Specifically, we first manually move the garment pieces to roughly align them with the body. During simulation, vertices between two ends of the seam gradually shrink until they completely pooled together. Figure~\ref{fig:ProceduralTexturing} (b) demonstrates the final draped garment.
For garment animation, we bind every vertex in the garment to the closest point in the body. Then, the skinning weight and blend shape of the body mesh are assigned to garment vertices. This makes our garment model easily integrated with existing skeletal pipelines with little computational overhead.


\noindent\textbf{Particle hair system.} 
Our method uses a prefabricated particle system to generate realistic hair on a head-shaped mesh. To achieve this, a template of hairstyle or facial hair $T_{hair}$ would attach to the vertices on the mesh which are marked with different areas $V_{hair}$ (including fringe, top, temporal bone, occipital bone, and bottom area). Designers draw multiple sets of guidelines $L_{guide}$ with varying shapes on different areas. Each set of guidelines comprises a collection of Bezier curves that are utilized to accurately constrain the flow of hair strands. Furthermore, the shape of the hair strands can be adjusted by length $P_{length}$ and curliness $P_{curliness}$. The entire hair system is composed of multiple sub-particle systems $\{V_{hair}, L_{guide}, P_{length}, P_{curliness}\}$.

\noindent\textbf{Accessories.}
We also add template accessories $T_{accessories}$ to our model, such as glasses, shoes, hats, and headphones. All accessories are pre-assembled on an SMPL-X template with a uniform body shape. These accessories can be transferred to models with the same topology, ensuring consistent deformation across all models in accordance with changes in the shape $\beta$ of SMPL-X. The corresponding bone weights of the human body model's vertices are transferred to the nearest accessories' vertices, ensuring that these accessories are correctly driven by the armature.

\noindent\textbf{Procedural texture.}
The procedural garment textures $T_{procedural} = \{T_{pattern}, T_{decals}, T_{bump}, P_{mapping}\}$ used in clothing are composed of multiple pre-set textures by alpha blending as demonstrated in Figure \ref{fig:ProceduralTexturing} (c). The pattern texture $T_{pattern}$ and decals texture $T_{decals}$ serve as layered masks to elevate the style and color, while the bump texture $T_{bump}$ functions as a height map, adding detailed normal information to the clothing's surface. The mapping parameters $P_{mapping}$ control the coordination of all the textures in the UV space. Besides, to procedurally create body mesh textures, we build a similar texture template SMPL-X, in which we pre-draw layered masks to separate different body features, such as skin, lips, and freckles, allowing for color adjustment and blending in specific regions as in Figure \ref{fig:SesneDressing}.

\begin{figure}[t]
  \centering
  \includegraphics[width=0.95\linewidth]{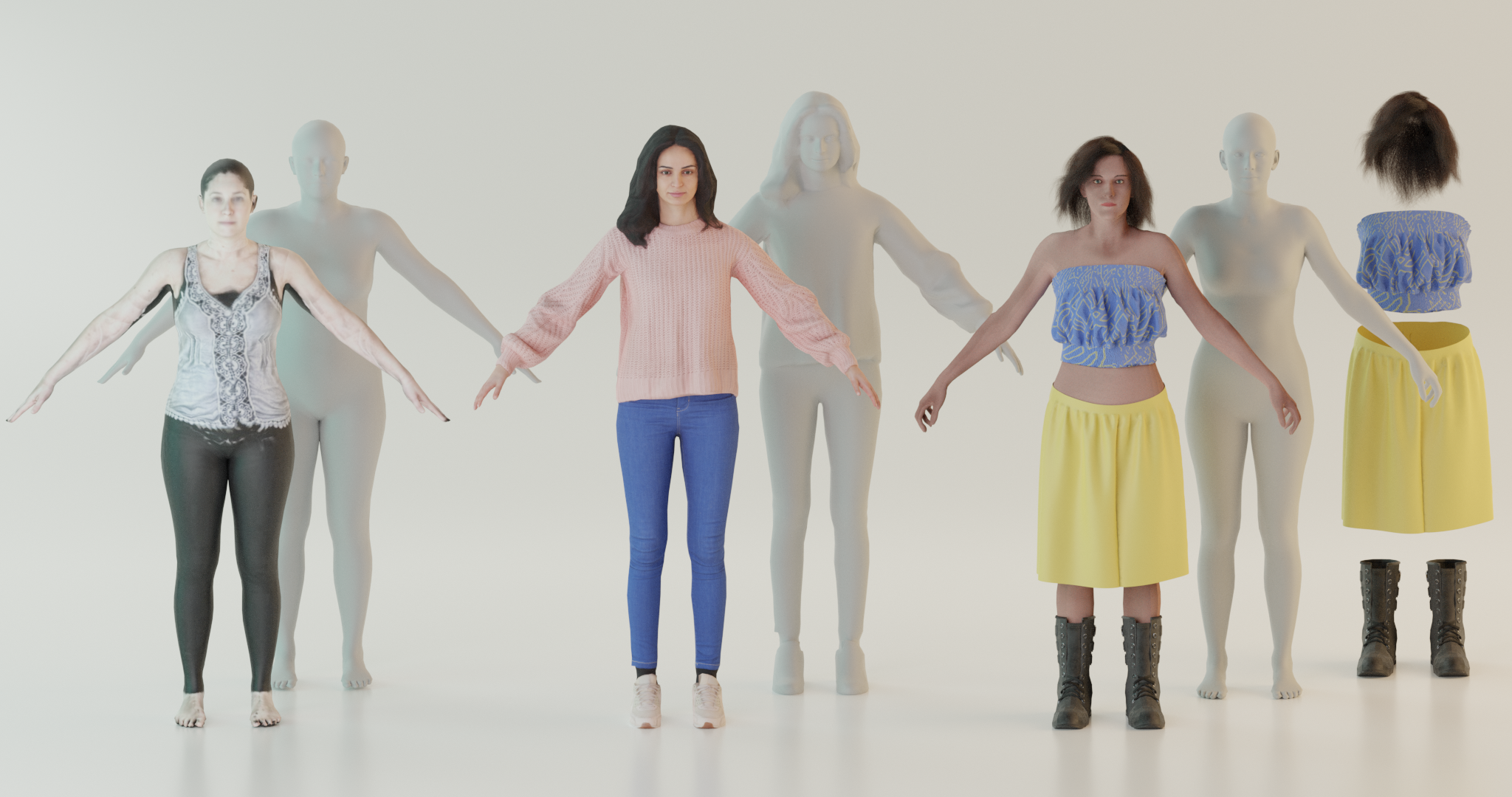}
  \centering
  {SURREAL~\cite{Varol2017LearningFS} ~~~~ RenderPeople~\cite{rp} ~~~~ \hmname}
  \caption{
A Demonstration of comparison between commonly used human models in existing synthetic datasets~\cite{Varol2017LearningFS,patel2021agora,bazavan2021hspace,cai2021playing} and our \hmname. We obtain high-quality models equal to RenderPeople \cite{rp}, both are much better than SURREAL \cite{Varol2017LearningFS}, and our model has the capability of scaling up easily by sampling various assets and body shapes.
}
  \label{fig:model-compare}
  \vspace{-0.2cm}
\end{figure}

Combining all the elements above, as shown in Figure~\ref{fig:model-compare}, we obtain human body models with the same high quality as RenderPeople~\cite{rp}.
Besides, it is challenging to expand the types of body shape and clothing available for real scans, but our model can be easily scaled by randomly sampling each component.

\subsection{Motion Retargeting}


\noindent\textbf{Retargeting of skeletal animations.}
%
Our motion retargeting module allows for the transfer of motion data from various sources, such as academic motion datasets, motion captures, and artist-crafted sources, to \hmname model skeletons. Despite variations in bone names, bone lengths, and rest pose bone rotations, the source skeletons are typically structurally similar to the target skeletons.

Pose frames in motion clips contain root translation $T(t) \in \mathbb{R}^3$ and rotations of each bone in the corresponding parent bone's space $\{R_{i}(t) \in \mathbb{R}^3 | i=0,1,...,n\}$. Following the forward kinematics (FK) manner, each bone's rotation relative to the model space can be calculated:
\begin{equation}
\widehat{R}_{i}(t) = \widehat{R}_{p(i)}(t) \cdot R_{i}(t) \
\end{equation}
where $\widehat{R}_i(t)$ is the rotation of $R_i(t)$ in model space at $t$ frame, and $p(i)$ indicates the parent of $i$. Specially, root bones have no parents so $\widehat{R}_{0}(t) = R_{0}(t)$.

To retarget motion from one skeleton to another, we assume the source and the target motion can drive the corresponding bones of both skeletons to the same rotation in model space.
We have T-pose frames of skeletons by manually posing them as T-posing, and treat T-pose as all motions' first frame pose. So motions relative to their T-pose frame can be easily obtained:
\begin{equation}
\widehat{R}_{i}(t) = \widehat{R}_{i}(0) \cdot \widehat{R}_{i\_src}(t) \cdot \widehat{R}^{-1}_{i\_src}(0).
\end{equation}
Considering skeletons have different bone lengths, which could result in ``sliding feet'' artifacts on target skeletons. We simply scale $T$ according to the ratio of pelvis bones' height to mitigate them:
$T(t) = \frac{H_{pelvis}}{H_{pelvis\_src}} \cdot T_{src}(t)$.
%
%

\noindent\textbf{SMPL-X Annotations.}
Considering \hmname body shapes are sampled from SMPL-X, the shape $\beta$ can be derived directly from the corresponding SMPL-X. Pose $\theta$ in the model space is calculated in the motion retargeting module. And in the scene composition module, models are placed in a 3D scene with camera space locations $T_c$ and rotations $R_c$. Pose $\theta_c$ in camera space is calculated by applying those world transformations to $\theta$. 
So SMPL-X annotations are constructed with $\{ \beta, \theta_c \}$. 

\noindent\textbf{SMPL Annotations.}
Although \hmname naturally provides accurate SMPL-X annotations, SMPL cannot be derived directly. Thus, we need to refit the SMPL parameters.
The optimization process consists of two steps. First, we fit the shape $\beta$ of all human models under the T-pose to its corresponding SMPL-X. Secondly, for each sequence, it is initialized with fitted $\beta$ and its original pose. we fix $\beta$ while fitting the body pose $\theta$ for SMPL.
%

\subsection{Scene Placement}

To place $N_p$ subjects in a large scene with $N_o$ objects, we primarily follow three principles: standing on the ground, avoiding human-object and human-human penetration. 
To prevent subjects from floating in the air, the root position of a subject should align with the ground height. A sequential decision-making approach is used to find a suitable position for each subject, \ie, placing one subject at a time.
An object is represented by $o_i=\{x_i, y_i,  l_i, w_i \} \in \mathbb{R}^4 $, where $\{x_i, y_i\}$ is the center of the object's axis-aligned bounding box projected onto the ground with length $l_i$ and width $w_i$. Different from static objects, to avoid collisions between moving subjects at any one time, a subject with specified body shape and motion is simplified to $q_i=\{x_i, y_i, l_i, w_i\} \in \mathbb{R}^4 $, which is the smallest axis-aligned box that envelops all bounding boxes across frames. 
\par

To avoid human-object penetration and potential human-human collision, the solution $p^*_i=\{x^*_i, y^*_i \}$ of a character with the shape of $\{l_i, w_i \}$ should satisfy:
\begin{equation}
I(\{x^*_i, y^*_i, l_i, w_i \}, o_j)=0, \ \ j=1,...,N_o
\label{posC1}
\end{equation}
\begin{equation}
I(\{x^*_i, y^*_i, l_i, w_i \},  q_k)=0, \ \ k=1,...,N_p,
\label{posC2}
\end{equation}
where $I(box_1,box_2)$ denotes the overlapping area between two boxes and $N_p$ is the number of subjects already placed in the scene.
Besides, distance constraint is used to prevent subjects from excessive dispersal. The problem is solved by grid search. Typically multiple solutions are available and we randomly sample one each time.

\subsection{Camera Placement}
\label{sec:toolchain_camera}
%
Once the positions of all subjects have been organized, the next task involves placing $N_c$ cameras in suitable locations. To ensure that cameras would not be placed inside any objects or subjects, the 3D version of Eq.~\eqref{posC1} and Eq.~\eqref{posC2} are applied.
In addition, we evaluate the suitability of a candidate camera by the following metrics: the distance from the camera to the subjects, the camera's pitch angle, and the degree of occlusion of each subject in the camera's view. 
\par
The distance denoted as $L$ from the mean position of all subjects $\bar{p}$ to the camera is restricted. 
$L_{max}$ is set to prevent an unreasonably small proportion of subjects in the image.
To control the visibility of subjects, $L_{min}$ is defined as
$\frac{\lambda}{sin(\alpha / 2)}\mathop{max}\limits_{i}||pv_i-\bar{p}||_2$, where $i=1,...,N_c$, 
$\alpha$ is the field of view of the camera, and $\lambda$ is a hyperparameter that determines the probability of all subjects being within the camera's view.
To estimate the degree of occlusion for each subject, rays are randomly and uniformly cast from the camera to each subject's body, and the percentage of blocked rays by other objects is determined.
%
More details can be found in the Sup. Mat.

\subsection{Rendering and Annotations}
Using a high-quality rendering pipeline in Unreal Engine 5, \name is rendered in multi-view with large 3D environments from the Unreal Marketplace for rich background information and dynamic lighting.
Leveraging the G-buffer~\cite{hargreaves2004deferred}, our system simultaneously generates photo-realistic RGB images and annotations, along with accurate ground truth for segmentation masks, optical flow, depth maps, normal maps, and other ground-truth labels.

\vspace{-5pt}
\section{\name Dataset}
\label{sec:dataset}



\subsection{Dataset Statistics}
%



\name comprises 10,000 unique \hmname models randomly created with different body shapes and genders. Each model is then combined with the following assets: (1) hairstyles $T_{hair}$ sampled from 45 particle hairs; (2) garments $G_{tmp}$ sampled from 68 clothing models containing multiple outfits; (3) procedural texture  $T_{procedural}$ generated by sampling $T_{pattern}$, $T_{decals}$, and $T_{bump}$ from 1,038 template textures, and color values are randomly sampled; (4) accessories $T_{accessories}$ sampled from 46 template assets.
%
To ensure the validity of the motions, we select a subset from AMASS~\cite{mahmood2019amass}. Based on the BABEL annotations~\cite{punnakkal2021babel}, we excluded interactive motions and non-ground motions(e.g., swim), as well as filtered out motions with a duration shorter than 2 seconds, leading to a subset with \ScaleMotion motions.
For each sequence, we randomly selected 4 human models, and each model is assigned a randomly chosen motion lasting between 2 and 10 seconds. For motions exceeding 2 seconds, we randomly extract a 2-second segment to ensure that each video clip has a length of 60 frames.
To enhance the diversity of viewpoints, 4 view positions are generated for each sequence, where each one is randomly sampled from the surface of semi-spheres with varying radii.
The system generates \ScaleSeq sequences and {\Scale}M images with annotations, utilizing 6 vast and realistic scenes created by professional artists. {\ScaleSMPL}M SMPL/SMPL-X annotations are provided, excluding highly occluded subjects.
More details can be found in the Sup. Mat.

\subsection{Human Pose and Shape Estimation}
With \ScaleSMPL million SMPLs, we leverage a pretrained regressor to extract 3D keypoints and then project them onto 2D space. Following standard practice in top-down human mesh recovery methods, we generate bounding boxes from the resulting 2D keypoints.

\subsection{Human Neural Rendering}
Given the flexibility of our pipeline, we render a total of 100 multi-view sequences with diverse motions and appearances for benchmarking human NeRFs. All sequences have a length of 300 frames with a resolution of $1024\times 1024$, where the motion sequence is randomly sampled from AMASS~\cite{mahmood2019amass}. Each sequence contains 8 views whose camera positions have uniformly distributed azimuth angles around the human body. We use RGB renderings, binary foreground masks, camera parameters, and SMPL parameters for the benchmark. Thanks to our layered design, we can offer accurate SMPL parameters for human NeRFs, which acts as an important prior for a majority of methods, while keeping the diversity in clothes and motions.

\begin{table}[t]
  \begin{center}
  \caption{Training popular baseline methods (image-based and video-based) with \name on 3DPW test set~\cite{von2018recovering}. R: standard real datasets. S: SynBody.}
  \label{tab:smpl_baseline}
  \vspace{-2mm}
  \resizebox{0.48\textwidth}{!}{
  \begin{tabular}{llccc}
    \toprule
    Method  & Datasets & MPJPE & PA-MPJPE & PVE \\    
    \midrule
    
    HMR \cite{kanazawa2018end}  & R & 112.30  & 67.50 & 141.92 \\

    HMR & R + S & \textbf{95.01}  & \textbf{57.62}  & \textbf{116.10} \\

    SPIN \cite{kolotouros2019learning} & R & 96.90  & 59.20  & 119.70 \\

    SPIN & R + S & \textbf{84.14}  & \textbf{53.67}  & \textbf{103.79} \\

    PARE \cite{kocabas2021pare} & R & 81.79  & 49.36  & 105.27 \\

    PARE & R + S & \textbf{78.98}  & \textbf{48.46}  & \textbf{103.86} \\

    VIBE \cite{kocabas2020vibe}  & R & 94.88 & 57.08  & 108.59 \\
    
    VIBE & R + S & \textbf{93.04}  & \textbf{57.00} & \textbf{107.23} \\
    
    \bottomrule
  \end{tabular}
  }
\end{center}
\vspace{-0.6cm}
\end{table}

\section{Experiment}
\label{sec:experiment}

In this section, we study the usefulness of \name for two popular research directions: human pose and shape estimation and human NeRF.

\subsection{Human Pose and Shape Estimation}
Estimating 3D humans represented by SMPL (body-only) and SMPL-X (body, hands, and face) parameters from monocular 2D input has gained substantial attention. \name emerges as a scalable and effective solution to tackle the scarcity of paired data in these fields. 

\noindent\textbf{Parametric Human Models.} 
SMPL~\cite{loper2015smpl} represents high-dimension human mesh $M(\theta,\ \beta)\ \in \mathbb{R}^{6890\times3}$ as low-dimension pose parameters $\theta\in \mathbb{R}^{72}$ and shape parameters $\beta\in \mathbb{R}^{10}$. In addition to SMPL, SMPL-X~\cite{pavlakos2019expressive} further consists of additional left and right hand pose parameters ($\theta_{lh}, \theta_{rh} \in \mathbb{R}^{15\times3}$), jaw joint rotation ($\theta_{lh} \in  \mathbb{R}^{3}$) and face expression parameters ($\phi_{f} \in \mathbb{R}^{10}$) for an expressive human representation. 

\noindent\textbf{Benchmarks.}
We conduct experiments on two mainstream benchmarks: 3DPW~\cite{von2018recovering} and AGORA~\cite{patel2021agora}. As a widely used benchmark for in-the-wild evaluation, 3DPW encompasses diverse data collected through a mobile phone camera, paired with SMPL annotations. As for AGORA, it is a recent synthetic dataset that features challenging scenes with person-person occlusion. AGORA provides both SMPL and SMPL-X annotations.

\begin{table}[t]
  \begin{center}
  \caption{Training popular baseline methods (image-based only as AGORA does not provide videos but static images only) with SynBody on AGORA validation set~\cite{patel2021agora}. R: standard real datasets. S: \name.}
  \label{tab:smpl_baseline_2}
  \vspace{-2mm}
  \resizebox{0.45\textwidth}{!}{
  \begin{tabular}{llccc}
    \toprule
    Method  & Datasets & MPJPE & PA-MPJPE & PVE \\    
    \midrule
    
    HMR~\cite{kanazawa2018end} & R & 226.71  & 87.72 & 248.35 \\

    HMR & R + S & \textbf{199.51}  & \textbf{77.97}  & \textbf{210.37} \\

    SPIN~\cite{kolotouros2019learning} & R & 212.91  & 79.76  & 217.88 \\

    SPIN & R + S & \textbf{196.81}  & \textbf{76.06}  & \textbf{205.83} \\

    PARE~\cite{kocabas2021pare} & R & 178.15  & 67.13  & 189.73 \\

    PARE & R + S &  \textbf{169.93} & \textbf{64.37}  & \textbf{179.81} \\
    
    \bottomrule
  \end{tabular}
  }
\end{center}
\end{table}

\noindent\textbf{Evaluation Metrics.}
To evaluate the quality of predicted parametric human models, we employ the standard metrics: \textit{MPJPE} (Mean Per Joint Position Error), which is calculated as the average $L_2$ distance over 3D keypoints regressed from parametric models; \textit{PA-MPJPE} (Procrustes Aligned Mean Per Joint Position Error), which is MPJPE with applying Procrustes Alignment~\cite{Gower1975} on the predicted keypoints to match the ground truth before error computation; \textit{PVE} (Per Vertex Error), which is the average $L_2$ distance between predicted and ground truth mesh vertices.

\subsubsection{SMPL Estimation}
\label{sec:exp:smpl}

\noindent\textbf{Methods.}
To gauge the usefulness of \name, we conduct experiments with four milestone works: HMR~\cite{kanazawa2018end}, SPIN~\cite{kolotouros2019learning}, and PARE~\cite{kocabas2021pare} are image-based methods and VIBE~\cite{kocabas2020vibe} is a video-based method. We follow the hyperparameter configurations provided in MMHuman3D~\cite{mmhuman3d}.

\noindent\textbf{Training Data.}
In the following experiments, we refer to the standard baskets of real datasets for SMPL estimation as "R". For HMR and SPIN, R consists of H36M\cite{h36m_pami}, MPI-INF-3DHP\cite{mehta2017monocular}, LSP\cite{johnson2010clustered}, LSPET \cite{Johnson2011LearningEH}, MPII\cite{andriluka20142d} and MSCOCO\cite{lin2014microsoft}, whereas in VIBE, R consists of MPI-INF-3DHP and InstaVariety\cite{humanMotionKZFM19}. "S" denotes SynBody data with 2.7M training instances with SMPL annotations.

\noindent\textbf{Main Results.}
We collate experiment results on 3DPW in Table~\ref{tab:smpl_baseline}. Each baseline is fine-tuned with a mixture of real datasets (R) and SynBody (S), leading to significant performance gains across different baseline models, even on strong baseline PARE~\cite{kocabas2021pare} (  2.81 mm and 1.41 mm improvements on MPJPE and PVE). Notably, \name can be used to train video-based methods such as VIBE as it contains video sequences instead of static images. 
Table \ref{tab:smpl_baseline_2} illustrates that training with \name exhibits remarkable improvements across methods, when compared with the baseline models in AGORA validation set.

\begin{figure}[t]
  \centering
  \includegraphics[width=\linewidth]{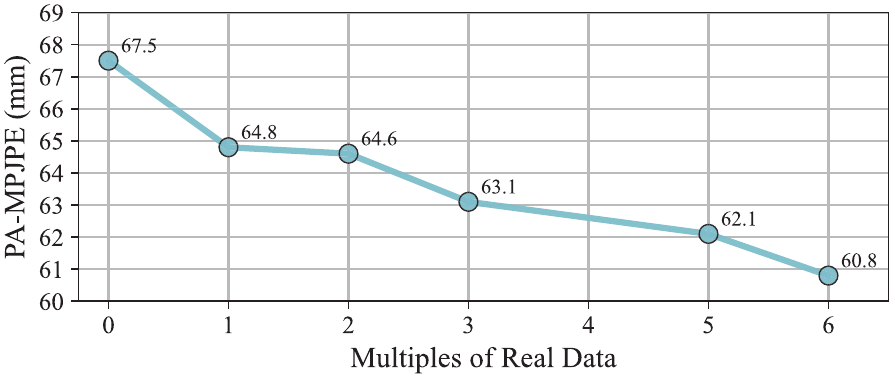}
  \vspace{-6mm}
  \caption{Impact of the amount of SynBody data. The horizontal axis represents the amount of SynBody data that is varied by multiples of real data. The baseline model is HMR and tested on the 3DPW test set.}
  \label{fig:multiplies}
\end{figure}
\begin{table}[t]
  \begin{center}
  \caption{Comparison between SynBody-100K and AGORA on 3DPW test set. ``A'' means AGORA, and ``S100K'' means SynBody-100K which is a subset of SynBody datasets with over 100K SMPL annotations. For a fair comparison, the total number of SMPL annotations is close to that of AGORA.}
  \label{tab:synbody_agora}
  \vspace{-2mm}
  \resizebox{0.47\textwidth}{!}{
  \begin{tabular}{llccc}
    \toprule
    Method & Datasets & MPJPE & PA-MPJPE & PVE \\    
    \midrule
    
    HMR & R + A & 101.61  & 57.85 & 123.82 \\
    
    HMR & R + S100K & \textbf{95.28}  & \textbf{57.68} & \textbf{119.18} \\
    
    SPIN & R + A & 88.44  & 54.97 & 110.35 \\

    SPIN & R + S100K & \textbf{85.52}  & \textbf{54.12} & \textbf{105.45} \\

    PARE & R + A & 85.34  & 48.39 & 109.77 \\
    
    PARE & R + S100K & \textbf{79.42}  & \textbf{47.80} & \textbf{102.45} \\
    
    \bottomrule
  \end{tabular}
  }
\end{center}
\end{table}
\begin{table}[t]
\centering
  \caption{Effectiveness of \hmname. R: real datasets (MSCOCO, MPI-INF-3DHP, Human3.6M, LSP, LSPET, MPII). S100K: downsampled SynBody with 100K instances. *: SURREAL-style.}
  \label{tab:LSMPL-X}
  \small
  \vspace{-2mm}
   \begin{tabular}{lcccc}
   \hline
    \toprule
    Method & Datasets & MPJPE & PA-MPJPE & PVE \\    
    \midrule
    PARE & R & 81.8 & 49.4 & 105.3 \\
    PARE & R+S100K* & 82.2 & 49.7 & 105.6 \\
    PARE & R+S100K & \textbf{79.4} & \textbf{47.8} & \textbf{102.5} \\
    \bottomrule
  \end{tabular}
  \vspace{2mm}
\end{table}
\begin{table}[t]
  \begin{center}
  \caption{SMPL-X estimation with OSX~\cite{lin2023one} as the baseline on AGORA validation set (AGORA-val)~\cite{patel2021agora} and 3DPW test set~\cite{lin2023one}. R: real datasets (MSCOCO, MPII, and Human3.6M). S: \name. AGORA-val uses PVE (mm) whereas 3DPW uses MPJPE (mm) and PA-MPJPE (mm). Note the original OSX uses the SMPL head for 3DPW, which we modify to the SMPL-X head.}
  \label{tab:smplx_baseline}
  \vspace{-2mm}  
  \resizebox{\linewidth}{!}{
  \begin{tabular}{lcccccc}
    \toprule
    \multirow{2}{*}{\textbf{Method}} & 
    \multirow{2}{*}{\textbf{Datasets}} &
    \multicolumn{3}{c}{\textbf{AGORA}} &
    \multicolumn{2}{c}{\textbf{3DPW (Body)}}  \\
    \cmidrule(lr){3-5}
    \cmidrule(lr){6-7}
    & & All & Hands & Face & MPJPE & PA-MPJPE
    \\
    \midrule
    OSX & R & 168.6 & 70.6 & 77.2 & 110.2 & 63.5 \\
    OSX & R+S600K & \textbf{155.8} & \textbf{63.5} & \textbf{72.0} & \textbf{92.2} & \textbf{59.3} \\
    \bottomrule
  \end{tabular}
  }
\end{center}
\end{table}

\begin{table*}[t]
  \begin{center}
  \caption{Benchmark of NeRF-based methods for 3D human neural rendering on \name.}
  \label{tab:humannerf}
  \small
  \vspace{-2mm}
  \begin{tabular}{lccccccccc}
    \toprule
    \multirow{2}{*}{\textbf{Method}}& \multicolumn{3}{c}{\textbf{Novel View}}  & \multicolumn{3}{c}{\textbf{Novel Pose}}  & \multicolumn{3}{c}{\textbf{Novel Identity}}\\ 
    & {PSNR\;$\uparrow$} & {SSIM\;$\uparrow$} & {LPIPS\;$\downarrow$} & {PSNR\;$\uparrow$} & {SSIM\;$\uparrow$} & {LPIPS\;$\downarrow$} & {PSNR\;$\uparrow$} & {SSIM\;$\uparrow$} & {LPIPS\;$\downarrow$}\\  
    \midrule
    NeRF~\cite{mildenhall_nerf_2020} & 19.39 & 0.862 & 0.162 & 19.61 & 0.824 & 0.201 & - & - & - \\
    NeuralBody~\cite{peng2021neural} &28.94&0.966&0.057&25.02&0.944&0.080&-&-&- \\
    HumanNeRF~\cite{weng_humannerf_2022_cvpr}&28.32&0.963&0.066&21.97&0.879&0.108&-&-&-\\
    AnimNeRF~\cite{peng2021animatable}&27.49&0.964&0.056&26.21&0.950&0.068&-&-&-\\
    NHP~\cite{kwon2021neural}&25.66&0.953&0.076&24.18&0.945&0.080&22.46&0.927&0.103\\

    \bottomrule
  \end{tabular}
\end{center}
\vspace{-3mm}
\end{table*}

\noindent\textbf{Impact of Data Scale.}
As our generation system can easily scale up the data size, we train HMR from scratch to study the influence of synthetic data scale.
Figure~\ref{fig:multiplies} demonstrates that adding more SynBody data generally leads to better performance. These experiments confirm that synthetic data is a valuable complement to real data and serves as a readily scalable training source to supplement the typically limited real data.

\noindent\textbf{Comparison with AGORA.}
To evaluate the quality of \name, we randomly sample a subset with 100K training instances, ``SynBody-100K'', which is comparable in size with the popular synthetic dataset AGORA. In Table~\ref{tab:synbody_agora}, we show that finetuning with \name outperforms that with AGORA across different baseline methods. The results highlight that \name demonstrates competitive traits as a training source.

\noindent\textbf{Effectiveness of \hmname.}
\hmname enables layered human modeling which cannot be achieved with SURREAL/RenderPeople as Figure~\ref{fig:model-compare}. We investigate the importance to synthesize data with these realistic considerations of \hmname in Table~\ref{tab:LSMPL-X}. We compare \name with SURREAL-style data that renders body and cloth texture on human mesh surfaces, without actual cloth geometry or hair. It is observed that SURREAL-style data leads to significant performance degradation.

\subsubsection{SMPL-X Estimation}
\label{sec:exp:smplx}

Also known as expressive human pose and shape estimation, SMPL-X estimation requires recovery of body, hands, and face parameters. 

\noindent\textbf{Method and Training Data.}
In Table~\ref{tab:smplx_baseline}, we conduct experiments with the recent SoTA, OSX~\cite{lin2023one}, as the base model. We compare the original baseline with training with a mixture of real datasets (``R" denotes COCO \cite{lin2014microsoft}, MPII \cite{andriluka20142d} and Human3.6M \cite{h36m_pami} with pseudo ground truth generated by NeuralAnnot \cite{moon2022neuralannot}) and \name (``S600K" here denotes a downsampled set of 600K instances as the real datasets are smaller for SMPL-X than that for SMPL estimation). Note that in the OSX paper, the values reported on 3DPW are obtained using a SMPL head. Here we standardize the output using the same SMPL-X head for both AGORA and 3DPW.
 
\noindent\textbf{Main Results.} 
We observe that adding SynBody in the training of OSX leads to a significant improvement in both overall estimation (more than 10 mm) and part-level (more than 5 mm for hands and face) estimation on AGORA. Moreover, in a fair comparison with the same model architecture, training with SynBody leads to 18 mm improvements in MPJPE on the 3DPW test set (here we follow the standard protocol to evaluate 14 key joints as 3DPW does not provide SMPL-X annotations). We speculate that it is difficult to obtain accurate SMPL-X annotations, which is mitigated by SynBody as it provides large-scale high-quality SMPL-X labels paired with images that are rendered with diverse backgrounds and lighting conditions, which are beneficial to training a high-performing expressive human parametric model recovery.

\subsection{Human NeRF}

In this section, we benchmark popular NeRF-based methods for 3D humans on \name, validating the effectiveness and great potential of our dataset in human neural rendering. Our benchmark is built upon three perspectives according to the purpose of synthesis tasks, which can be categorized into novel view, novel pose, and novel identity.

\noindent\textbf{Methods for Benchmark.}
We benchmark five methods in total, including the vanilla NeRF~\cite{mildenhall_nerf_2020}, NeuralBody~\cite{peng2021neural} and HumanNeRF~\cite{weng_humannerf_2022_cvpr} for novel view synthesis, AnimNeRF~\cite{peng2021animatable} for novel pose synthesis, NHP~\cite{kwon2021neural} for generalizable human NeRF (novel identity synthesis). Except for NHP, all methods are trained in a person-specific manner, taking 4 views of the first 250 frames for training and the rest views and frames for evaluations.

\noindent\textbf{Evaluation Protocols.}
We follow~\cite{mildenhall_nerf_2020, weng_humannerf_2022_cvpr} to evaluate all methods using three standard metrics: Peak Signal-to-Noise Ratio (PSNR), Structural SIMilarity index (SSIM), and Learned Perceptual Image Patch Similarity (LPIPS~\cite{zhang2018unreasonable}). To reach a consensus among all methods, all metrics are computed over the whole image with a black background. 

\begin{figure}[t]
    \centering
    \includegraphics[width=\linewidth]{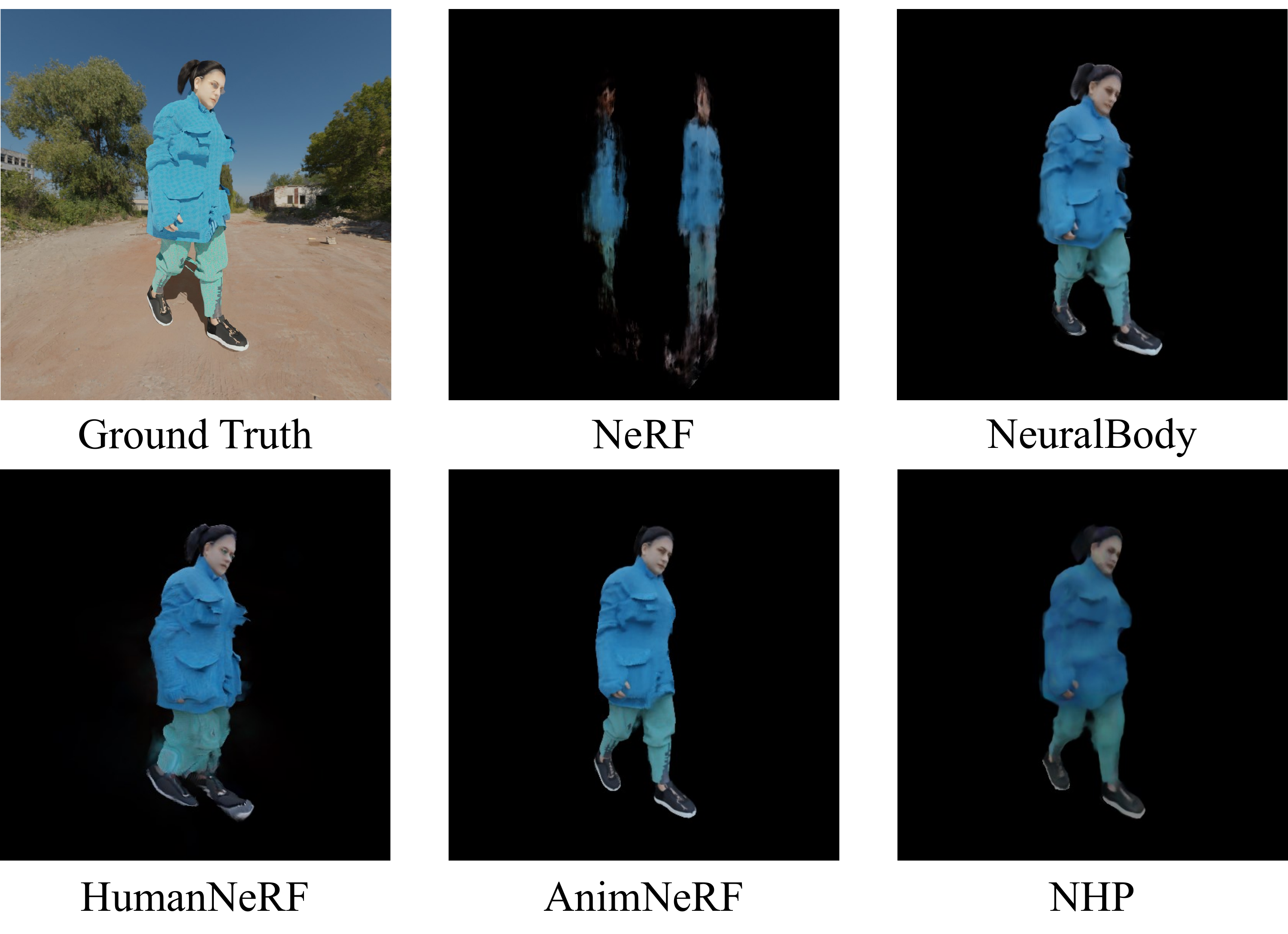}
    \vspace{-6mm}
    \caption{Novel view synthesis of different human NeRF methods on \name.}
    \label{fig:humannerf}
    \vspace{-0.2cm}
\end{figure}

\noindent\textbf{Main Results.}
Benchmark results are reported in Table \ref{tab:humannerf}. We observe that all methods achieve comparable performances as real human data on \name. Models (NeuralBody and AnimNeRF) which rely on accurate SMPL estimation and blending weights perform better on novel poses. We attribute it to our layer-wise design which offers ground truth SMPL parameters for human NeRF training. Besides, we present visualization results in Figure~\ref{fig:humannerf}, observing that the diverse motions and appearances, as well as loose garments in \name, pose further challenges in the field of neural rendering of 3D humans.

\vspace{-5pt}
\section{Conclusion}
\label{sec:conclusion}
We present \name, a large-scale synthetic dataset that features a substantial number of subjects and high-quality 3D annotations. At the core is a clothed human model with multiple layers of representation.
Our experiments demonstrate the effectiveness of SynBody on both human mesh recovery and human NeRF.
Future research can leverage \name for developing and evaluating methods to predict body and cloth simultaneously.
Furthermore, the high controllability of the synthetic dataset offers ample opportunities for further improvements, such as the incorporation of contact labels for human-scene interaction.

\noindent\textbf{Societal Impacts.}
Even though \name is a synthetic dataset, the assets utilized in its creation might not be well-balanced. Although hairstyles and skin colors are chosen at random, they are designed to avoid racial biases. Yet, other elements, such as body shapes and clothing, might not be as balanced, posing a potential for bias in the resultant human models.

\noindent\textbf{Acknowledgement.}
This study is supported by the National Research Foundation, Singapore under its AI Singapore Programme (AISG Award No: AISG2-PhD-2021-08-019), the Ministry of Education, Singapore, under its MOE AcRF Tier 2 (MOE-T2EP20221-0012), NTU NAP, and under the RIE2020 Industry Alignment Fund – Industry Collaboration Projects (IAF-ICP) Funding Initiative, as well as cash and in-kind contribution from the industry partner(s).

{\small
\bibliographystyle{ieee_fullname}
\bibliography{egbib}
}

\clearpage
\appendix
{\noindent\Large\textbf{Supplementary Material}}

\section{Details of Synthetic Data Generation}

\subsection{Construction of AMASS subset}

To ensure the validity of the motions, we selected a subset from AMASS~\cite{mahmood2019amass}. Following the BABEL annotations~\cite{punnakkal2021babel}, we excluded interactive motions, non-ground motions, and motions with a duration of less than 2 seconds. The specific categories excluded were: ``unknown'', ``interact with/use object'', ``touching body part'', ``exercise/training'', ``move up/down incline'', ``sit'', ``touch object'', ``touching face'', ``swim'' and `fall.' The resulting subset comprised \ScaleMotion motion sequences, each lasting more than 2 seconds.

\subsection{Details of Camera Placemeent}

\begin{figure}[t]
  \includegraphics[width=0.9\linewidth]{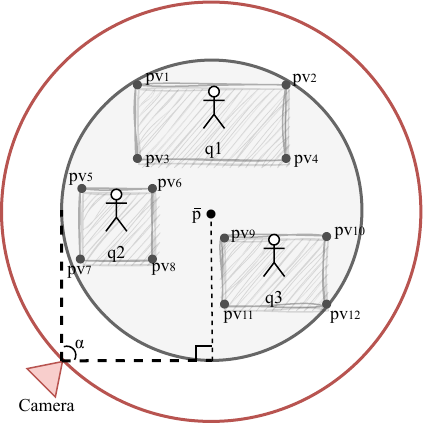}
  \caption{Illustration of camera placement.}
  \label{fig:supp_camera}
  \vspace{-0.2cm}
\end{figure}

Figure~\ref{fig:supp_camera} illustrates the process of calculating the camera's minimum distance to the subjects' center.
The shaded box represents the envelop box of a subject across frames, while the gray circle ensures that it encompasses all subjects. Therefore, the camera should ensure that the circle is within its field of view, which requires the distance to be greater than $L_{min}$, the radius of the red circle.
As stated in the main text, $L_{min} = 
\frac{\lambda}{sin(\alpha / 2)}\mathop{max}\limits_{i}||pv_{i}-\bar{p}||_2$. 
In this equation, $pv$ represents the position of the vertex of the envelope box of the subject, with $i=1,...,N_v$, where $N_v$ is the total number of vertices.
Additionally, cameras that have a pitch angle outside of a predefined range which is $[-5^\circ, 30^\circ]$ will be excluded from consideration,
$L_{max}$ is set to $10$ meters for preventing an unreasonably small proportion of subjects in the image.

\section{Data examples of \name}

\subsection{Image examples}

\begin{figure*}[t]
  \includegraphics[width=\linewidth]{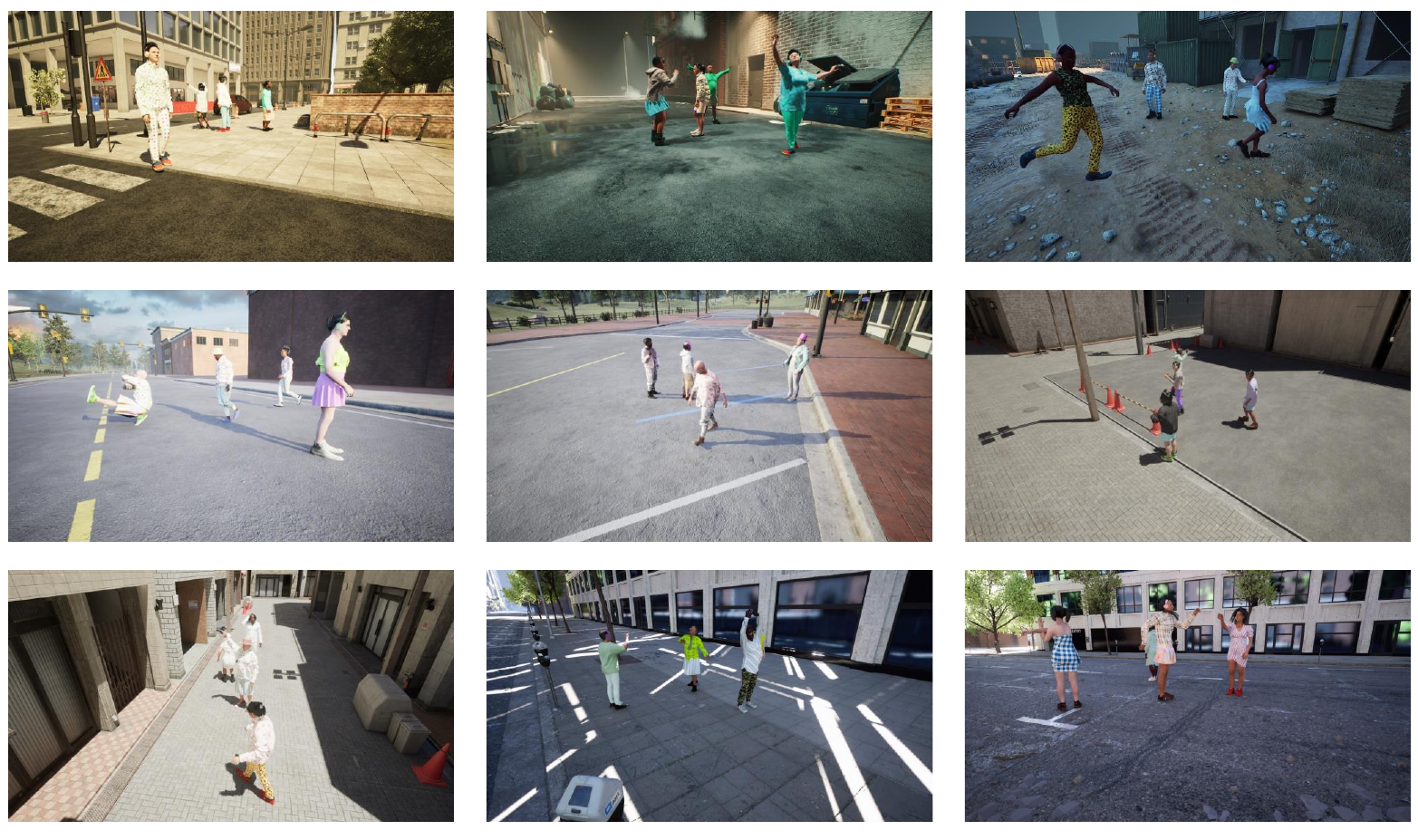}
  \caption{Illustration of synthetic images. \name features subjects with a variety of appearances and poses. These subjects are captured from various camera angles, set against diverse, realistic backgrounds, and illuminated under different lighting conditions. These considerations are critical to the usefulness of \name across various tasks.}
  \label{fig:supp_images}
\end{figure*}

In the main paper, we show 1.2M images (2.7M instances in 27K sequences) containing subjects of neutral gender. However, \name also contains rendered images with three genders (neutral, female, male) that add up to $\sim$1.6M images (6M instances in 38K sequences) in the grand total. We show image examples in Figure~\ref{fig:supp_images}.

\subsection{Annotation examples}

\begin{figure*}[t]
  \includegraphics[width=\linewidth]{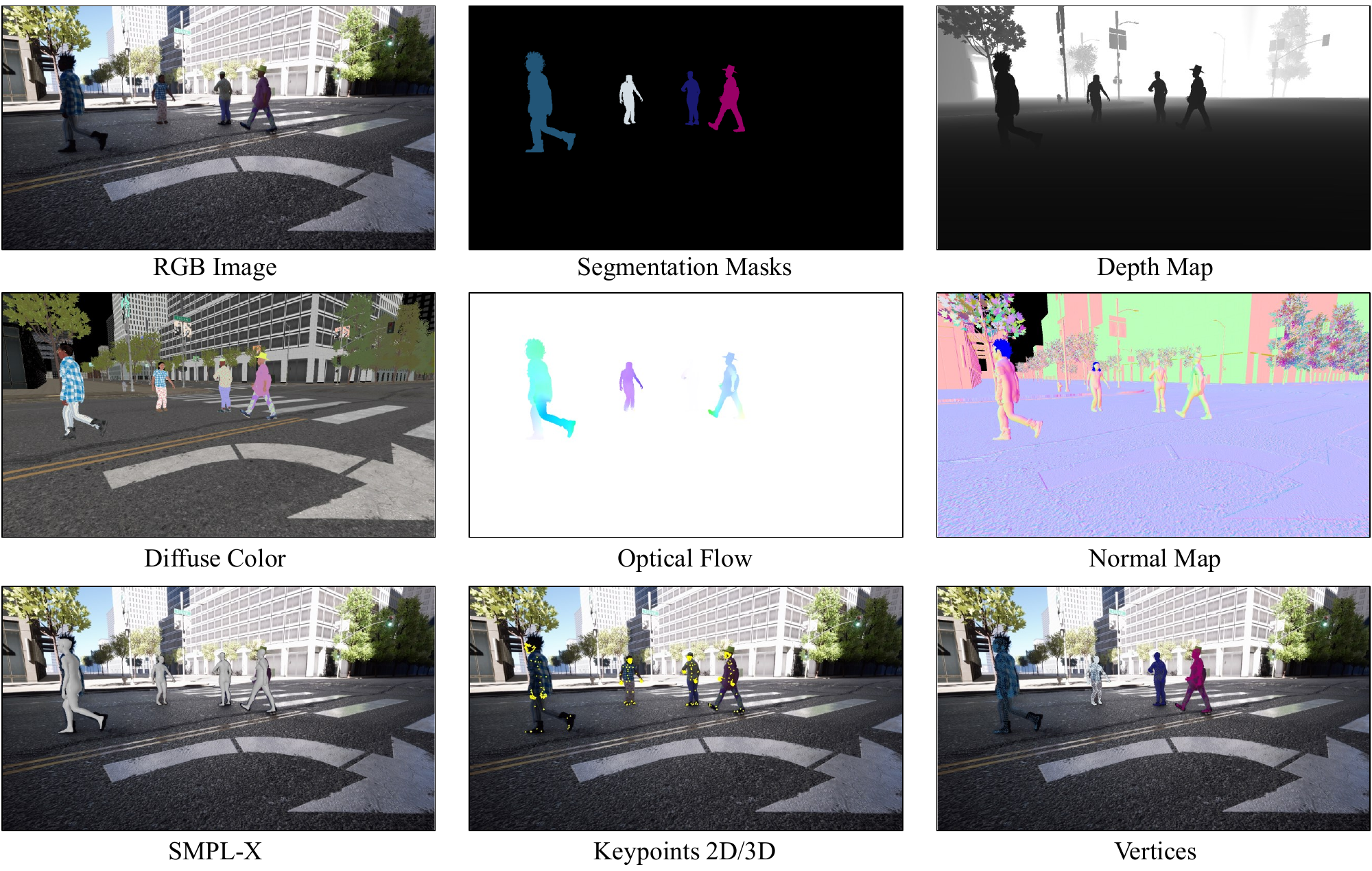}
  \caption{Illustration of annotations. \name provides accurate annotations paired with RGB images. Therefore, \name can support a myriad of human-related tasks in perception and reconstruction. }
  \label{fig:supp_annotation}
\end{figure*}

\name features accurate and diverse annotations that support various human perception and reconstruction tasks. In Figure~\ref{fig:supp_annotation}, we show an RGB image with paired labels such as segmentation masks, keypoints, normal map, and SMPL-X. We highlight that some labels are expensive to obtain in real life, making \name a promising alternative for scaling up training data.

\section{Asset examples of \name}

\name utilizes a wide range of 3D assets in the rendering. These assets enhance the realism and diversity of generated images.

\subsection{Scenes}

\begin{figure*}[t]
  \centering
  \includegraphics[width=0.8\linewidth]{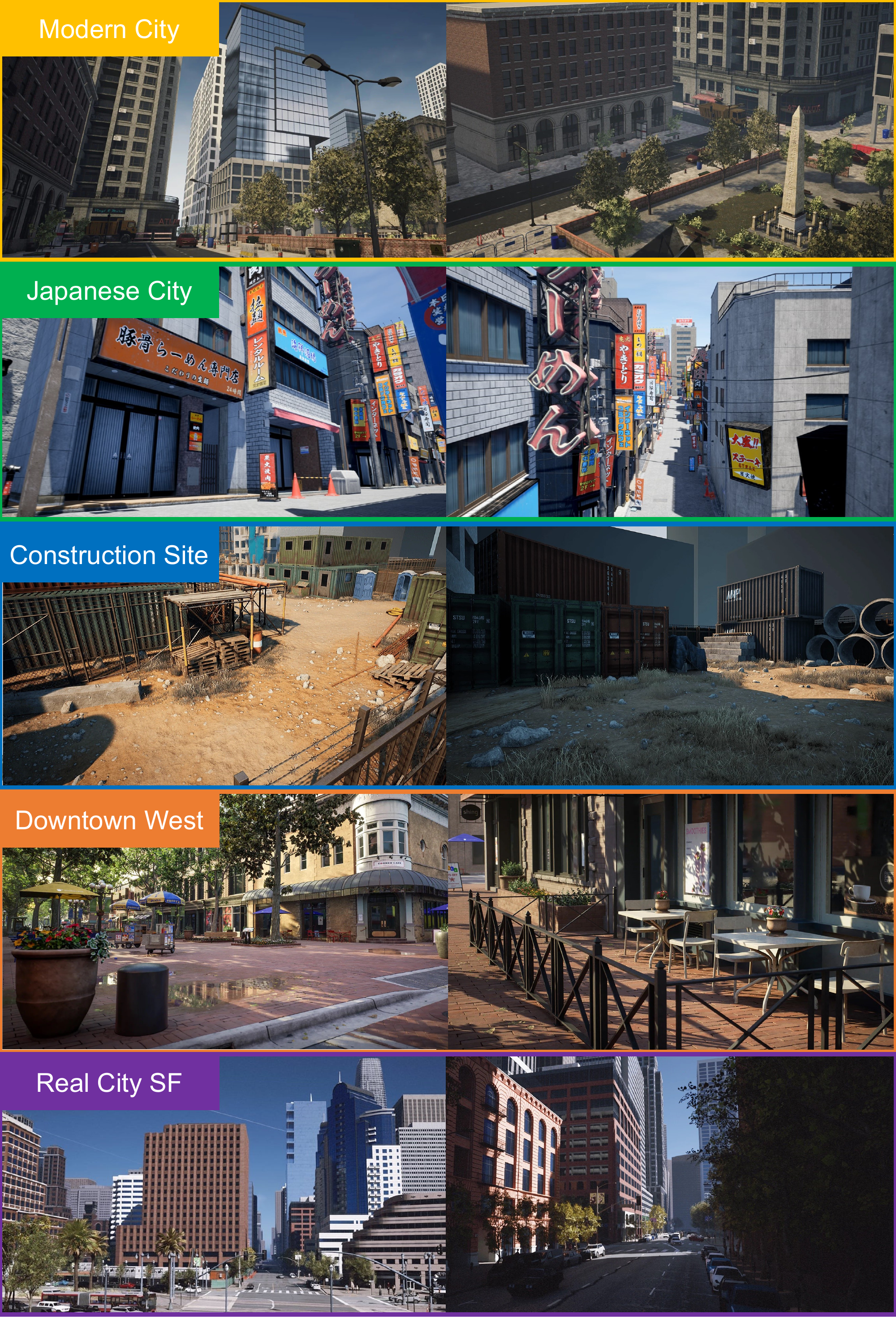}
  \caption{Illustration of scenes. We utilize high-quality, diverse city-scale scene models in rendering our images.}
  \label{fig:supp_scene}
\end{figure*}

In Figure~\ref{fig:supp_scene}, we place our virtual subjects in expansive, meticulously crafted 3D scenes. These environments are not only vast in scale but also emulate lifelike atmospheres, capturing a myriad of architectural designs from various cultures. We argue that the intrinsic diversity and high-fidelity quality of these backgrounds not only enhance the visual appeal but also play a pivotal role in potentially mitigating the synthetic-real domain gap.

\subsection{Hairstyles, Clothes, and Accessories}

\begin{figure*}[t]
  \centering
  \includegraphics[width=\linewidth]{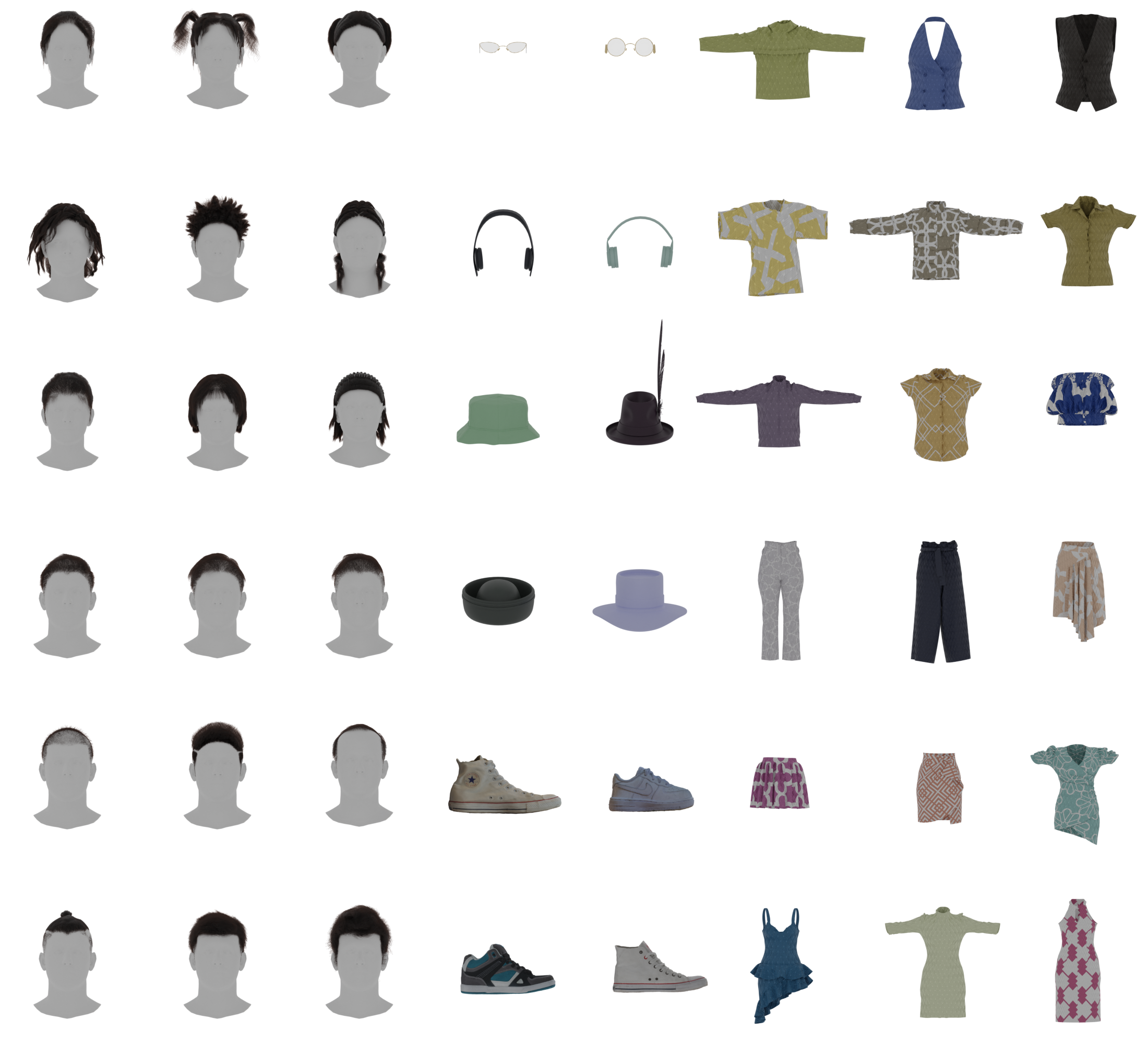}
  \caption{Illustration of Assets used in \hmname. \hmname enables a layered human modeling that encompasses a wide range of hairstyles, accessories (such as hats and shoes), and clothes of different types, dimensions, and textures. }
  \label{fig:supp_cloth}
\end{figure*}

One of the standout features of \hmname is the extensive collection of appearance elements beyond the naked human body mesh. In Figure~\ref{fig:supp_cloth}, we demonstrate a vast repository of diverse hairstyles, clothing (with procedural textures, and accessories (such as glasses, shoes, hats, and headphones). These elements enhance the depth of detail and customization available in our \hmname: a comprehensive layered human representation.

\end{document}